\documentclass{article}
\usepackage[T1]{fontenc}
\usepackage[letterpaper, margin=1.2in, top=1.2in, bottom=1.2in]{geometry}
\usepackage{times}
\usepackage{helvet}
\usepackage{microtype}

\usepackage{amsmath}
\usepackage{amsfonts}
\usepackage{amssymb}
\usepackage{graphicx}
\usepackage{natbib}
\usepackage{enumitem}
\usepackage{pifont}
\usepackage{fontawesome5}
\usepackage{nicefrac}
\usepackage[table]{xcolor}

\usepackage{booktabs}
\usepackage{caption}
\usepackage{subcaption}
\usepackage{multirow}
\usepackage{makecell}
\usepackage{arydshln}
\usepackage{longtable}
\usepackage{tabularx}
\usepackage{array}
\usepackage{colortbl}

\usepackage{algorithm}
\usepackage{algpseudocode}
\usepackage{fvextra}

\usepackage{tcolorbox}
\tcbuselibrary{skins, breakable}
\usepackage[framemethod=tikz]{mdframed}
\usepackage{titlesec}
\usepackage{authblk}
\usepackage{fancyhdr}
\usepackage{tikz}
\usepackage{wrapfig}

\usepackage{hyperref}
\usepackage{cleveref}
\Crefname{figure}{Figure}{Figures}
\Crefname{table}{Table}{Tables}
\Crefname{section}{Section}{Sections}
\Crefname{equation}{Equation}{Equations}
\Crefname{algorithm}{Algorithm}{Algorithms}
\Crefname{appendix}{Appendix}{Appendix}

\definecolor{darkblue}{RGB}{0, 51, 102}
\definecolor{lightbluebg}{RGB}{245, 248, 252}
\definecolor{blueframe}{RGB}{90, 150, 200}
\definecolor{dividercolor}{RGB}{200, 210, 220}
\definecolor{yellowtext}{RGB}{68,132,243}
\definecolor{yellowred}{RGB}{50,167,82}
\definecolor{yellowblue}{RGB}{251,191,5}
\definecolor{darkgreen}{rgb}{0,0.4,0}
\definecolor{maroon}{HTML}{A00000}
\definecolor{gray}{rgb}{0.5, 0.5, 0.5}
\definecolor{chocolate}{HTML}{D2691E}
\definecolor{indigo}{HTML}{4B0082}
\definecolor{violet}{HTML}{4B2E83}
\definecolor{lightgreen}{HTML}{E0FBE0}
\definecolor{lightred}{HTML}{FBE0E0}
\definecolor{cadmiumgreen}{rgb}{0.0, 0.42, 0.24}
\definecolor{forestgreen}{rgb}{0.13, 0.55, 0.13}
\definecolor{lightgray}{rgb}{0.9, 0.9, 0.9}
\definecolor{exp_table_blue}{HTML}{DAECED}

\definecolor{myblue}{rgb}{0.2, 0.4, 0.8}
\definecolor{mygreen}{rgb}{0.1, 0.5, 0.1}
\definecolor{evidencered}{RGB}{230, 30, 30}
\definecolor{timeblue}{RGB}{40, 90, 170}
\definecolor{backendblue}{HTML}{5898D8}

\hypersetup{
    colorlinks=true,
    linkcolor=darkblue,
    citecolor=darkgreen,
    urlcolor=darkblue
}

\titleformat{\section}{\sffamily\Large\bfseries\color{darkblue}}{\thesection}{1em}{}
\titleformat{\subsection}{\sffamily\large\bfseries\color{darkblue}}{\thesubsection}{1em}{}
\titleformat{\subsubsection}{\sffamily\normalsize\bfseries\color{darkblue}}{\thesubsubsection}{1em}{}
\titleformat{\paragraph}[runin]{\sffamily\normalsize\bfseries}{}{0em}{}[]
\titlespacing{\paragraph}{0pt}{0.5ex plus 0.2ex minus 0.1ex}{0.5em}

\setlist[enumerate,itemize]{topsep=0pt, itemsep=0pt, leftmargin=*, after=\leavevmode}
\setlist[enumerate,1]{label=(\arabic*)}

\newcommand{\cmarkT}{\textcolor{green!70!black}{\ding{51}}}
\newcommand{\cmarkF}{\textcolor{red}{\ding{55}}}

\DeclareRobustCommand{\evidencebox}{%
    \tikz[baseline=-0.5ex]\draw[evidencered, line width=0.9pt, rounded corners=1.5pt]
        (0,0) rectangle (0.95em, 0.6em);}
\DeclareRobustCommand{\timebox}{%
    \tikz[baseline=-0.5ex]\draw[timeblue, line width=0.9pt, rounded corners=1.5pt]
        (0,0) rectangle (0.95em, 0.6em);}

\DefineVerbatimEnvironment{VerbatimWrap}{Verbatim}{
    breaklines=true,
    breakindent=0pt,
    breaksymbol={},
    fontsize=\scriptsize,
}

\mdfdefinestyle{customstyle}{
    linecolor=cyan!70,
    linewidth=3pt,
    innerleftmargin=5pt,
    topline=false,
    rightline=false,
    bottomline=false,
    leftline=true,
    innerrightmargin=5pt,
    innertopmargin=5pt,
    innerbottommargin=5pt,
    backgroundcolor=cyan!8,
}

\newtcolorbox{prompt}[1]{
    colback=lightbluebg!30!white,
    colframe=blueframe,
    breakable,
    title=\textit{#1}
}

\makeatletter
\def\@title{}
\def\@abstract{}
\def\@keywords{}
\def\@codelink{}
\def\@datasetlink{}
\def\@projectlink{}
\def\@footnotemarkers{}

\renewcommand{\title}[1]{\def\@title{#1}}
\renewcommand{\abstract}[1]{\def\@abstract{#1}}
\newcommand{\keywords}[1]{\def\@keywords{#1}}
\newcommand{\codelink}[1]{\def\@codelink{#1}}
\newcommand{\datasetlink}[1]{\def\@datasetlink{#1}}
\newcommand{\projectlink}[1]{\def\@projectlink{#1}}
\newcommand{\footnotemarkers}[1]{\def\@footnotemarkers{#1}}

\fancypagestyle{firstpage}{
  \fancyhf{}
  \fancyfoot[L]{\rule{0.333\textwidth}{0.4pt}\\[2pt]\footnotesize \@footnotemarkers}
  
}

\renewcommand{\maketitle}{%
  \thispagestyle{firstpage}%
  \begin{tcolorbox}[
    breakable,
    colback=lightbluebg,
    colframe=blueframe,
    arc=3mm,
    boxrule=0.5pt,
    left=12pt,
    right=12pt,
    top=12pt,
    bottom=12pt,
    width=\textwidth,
    boxsep=5pt
  ]
    \noindent
    {\sffamily\Large\bfseries\color{darkblue}\@title\par}
    \vspace{0.6em}
    \noindent
    \parbox{\textwidth}{\@author}\par
    \vspace{1.2ex}
    {\color{dividercolor}\rule{\linewidth}{0.5pt}}\par
    \vspace{1.2ex}
    \ifx\@abstract\@empty\else
      \noindent\parbox{\textwidth}{%
        {\sffamily\bfseries Abstract:}\quad \@abstract%
      }\par
      \vspace{1ex}
    \fi
    \ifx\@keywords\@empty\else
      \noindent\parbox{\textwidth}{%
        {\sffamily\bfseries Keywords:}\quad \@keywords%
      }\par
      \vspace{1.5ex}
    \fi
    \ifx\@codelink\@empty
      \ifx\@datasetlink\@empty
        \ifx\@projectlink\@empty
        \else
          \vspace{-1.1em}
          {\color{dividercolor}\rule{\linewidth}{0.5pt}}
          \vspace{-0.6em}
          \centering
          \sffamily\small
          \href{\@projectlink}{\faGlobe\ Website}
          \par
        \fi
      \else
        \vspace{-1.1em}
        {\color{dividercolor}\rule{\linewidth}{0.5pt}}
        \vspace{-0.6em}
        \centering
        \sffamily\small
        \ifx\@projectlink\@empty\else
          \href{\@projectlink}{\faGlobe\ Website}\hspace{1.5em}
        \fi
        \href{\@datasetlink}{\faDatabase\ Benchmark}
        \par
      \fi
    \else
      \vspace{-1.1em}
      {\color{dividercolor}\rule{\linewidth}{0.5pt}}
      \vspace{-0.6em}
      \centering
      \sffamily\small
      \ifx\@projectlink\@empty\else
        \href{\@projectlink}{\faGlobe\ Website}\hspace{1.5em}%
      \fi
      \href{\@codelink}{\faGithub\ Code}%
      \ifx\@datasetlink\@empty\else
        \hspace{1.5em}\href{\@datasetlink}{\faDatabase\ Benchmark}%
      \fi
      \par
    \fi
  \end{tcolorbox}
}
\makeatother

\title{StreamArena: Toward Continuous, Interactive, and Long-Horizon Agentic Streaming Video Understanding}

\author[1,2,$*$,$\ddagger$]{Xichen Zhang}
\author[2,$*$]{Guankai Li}
\author[3]{Yinghao Zhu}
\author[2]{Shijian Wang}
\author[4]{Sitong Wu}
\author[4]{Shaozuo Yu}
\author[1]{Meng Chu}
\author[2]{Yuan Lu}
\author[1,$\dagger$]{Jiaya Jia}

\affil[1]{The Hong Kong University of Science and Technology}
\affil[2]{Xiaohongshu Inc.}
\affil[3]{The University of Hong Kong}
\affil[4]{The Chinese University of Hong Kong}

\abstract{%
Deploying autonomous multimodal agents in continuous, real-world environments requires them to ingest unbounded audio-visual streams and maintain hour-scale memory. However, current evaluations predominantly rely on brief clips and multiple-choice formats. This design allows minimal baselines that process only the last four frames to match or surpass complex streaming models, while answer options also expose language shortcuts. We introduce \textit{StreamArena}, a benchmark for hour-scale, interactive streaming video understanding. StreamArena contains 243 full-length videos averaging 88.8 minutes and 3,646 rigorously annotated, open-ended question-answer pairs that evaluate real-time perception, historical retrospection, proactive interaction, and multimodal tool utilization. Evaluation across diverse systems exposes a tension between continuous interaction and long-horizon multimodal comprehension. Methods that retain only recent frames cannot recover distant events, methods that convert past observations into text lose visual evidence, and methods that repeatedly compress visual memory struggle to preserve fine-grained details over time. We address this tension with \textit{StreamMind}, a two-tier architecture that assigns latency-critical interaction and proactive monitoring to independently scheduled frontend workers, while backend workers asynchronously construct persistent multimodal memory and perform historical recall and external search. StreamMind outperforms existing streaming baselines across all four capabilities and reduces query-to-answer latency by reusing persistent state.
}

\footnotemarkers{%
$^*$Equal contribution. \quad
$^\ddagger$Work done during an internship at Xiaohongshu Inc.\\
$^\dagger$Corresponding author: Jiaya Jia (\texttt{jia@cse.ust.hk}).}


\projectlink{https://hkuzxc.github.io/StreamArena_web/}
\codelink{https://github.com/JIA-Lab-research/StreamArena}
\datasetlink{https://huggingface.co/datasets/hkuzxc/StreamArena}

\begin{document}

\maketitle

\section{Introduction}
\label{sec:introduction}

% =============================================================
% Table: Comparison with existing video understanding benchmarks
% Columns (12): Benchmark | Venue | Anno. | Ans. | Q/Vid | Dur 
%              | Long | Str. | Omni | MT | Pro. | Tool
%
% Grouping:
%   (A) Offline long / omni-modal video benchmarks
%   (B) Online / streaming video benchmarks
%   (C) Proactive / duplex streaming benchmarks
%   (D) Ours
%
% Inclusion policy (per author decision):
%   - Only officially accepted works are included.
%
% Notes:
%   - Anno.: Manual = fully human-authored; Hybrid = LLM draft + human curation.
%   - Ans.: Open = free-form; MC = multiple-choice; Mixed = both.
%   - Q/Vid = mean questions per video; Dur = mean duration in minutes.
%   - Capability columns: Long ($\geq$10 min), Str. (causal / prefix-only),
%     Omni (video+audio), MT (multi-turn dialogue), Pro. (proactive
%     when-to-respond), Tool (external tool invocation).
% =============================================================
\begin{table*}[t]
\centering
\footnotesize
\setlength{\tabcolsep}{5.5pt}
\renewcommand{\arraystretch}{1.05}
% \vspace{-2mm}
\scalebox{0.94}{%
\begin{tabular}{@{}l l c c r r c c c c c c@{}}
\toprule
\textbf{Benchmark} & \textbf{Venue}
 & \textbf{Anno.} & \textbf{Ans.}
 & \textbf{Q/Vid} & \textbf{Dur}
 & \textbf{Long} & \textbf{Str.} & \textbf{Omni}
 & \textbf{MT} & \textbf{Pro.} & \textbf{Tool} \\
\midrule
\multicolumn{12}{l}{\textit{(A) Offline long / omni-modal video benchmarks}} \\
Video-MME             & CVPR'25    & Manual  & MC    &  3 & 17.0 & \cmarkT & \cmarkF & \cmarkT & \cmarkF & \cmarkF & \cmarkF \\
LongVideoBench        & NeurIPS'24 & Manual  & MC    &  2 & 7.9  & \cmarkT & \cmarkF & \cmarkF & \cmarkF & \cmarkF & \cmarkF \\
MLVU                  & CVPR'25    & Manual  & Mixed &  2 & 15.5 & \cmarkT & \cmarkF & \cmarkF & \cmarkF & \cmarkF & \cmarkF \\
\midrule
\multicolumn{12}{l}{\textit{(B) Online / streaming video benchmarks}} \\
StreamingBench$^{\ddagger}$ & ICLR'25    & Hybrid  & MC    &  5 &  9.7 & \cmarkF & \cmarkT & \cmarkT & \cmarkT & \cmarkT$^{\mathsection}$ & \cmarkF \\
OVO-Bench$^{\ddagger}$  & CVPR'25    & Hybrid  & Mixed &  4 &  3.5 & \cmarkF & \cmarkT & \cmarkF & \cmarkF & \cmarkT$^{\mathsection}$ & \cmarkF \\
OVBench               & CVPR'25    & Hybrid  & MC    & 11 &  5.5 & \cmarkF & \cmarkT & \cmarkF & \cmarkT & \cmarkF & \cmarkF \\
RTV-Bench             & NeurIPS'25 & Hybrid  & MC    &  8 & 18.2  & \cmarkF & \cmarkT & \cmarkF & \cmarkF & \cmarkF & \cmarkF \\
OST-Bench             & NeurIPS'25 & Hybrid  & Mixed &  7 & N/A  & \cmarkF & \cmarkT & \cmarkF & \cmarkF & \cmarkF & \cmarkF \\
\midrule
\multicolumn{12}{l}{\textit{(C) Proactive / duplex streaming benchmarks}} \\
OmniMMI               & CVPR'25    & Manual  & Mixed &  2 &  5.4 & \cmarkF & \cmarkT & \cmarkT & \cmarkT & \cmarkT & \cmarkF \\
QIVD (Qualcomm)       & ICLR'26    & Manual  & Open  &  1 & 0.09 & \cmarkF & \cmarkT & \cmarkT & \cmarkF & \cmarkT & \cmarkF \\
\midrule
\rowcolor{gray!12}
\textbf{Ours}         & ---      & \textbf{Manual} & \textbf{Open}
                      & \textbf{$\sim$15} & \textbf{88.8}
                      & \cmarkT & \cmarkT & \cmarkT & \cmarkT & \cmarkT & \cmarkT \\
\bottomrule
\end{tabular}
}
\caption{Properties of video understanding benchmarks. Anno. and Ans. denote annotation and answer formats; Q/Vid and Dur denote mean questions per video and mean duration in minutes. Capability columns cover long videos, causal streaming, joint video and audio, multi-turn dialogue, proactive triggering, and external tools. $^{\ddagger}$ marks documented recency shortcuts~\cite{shen2026simple}; $^{\mathsection}$ marks separate queries before and at the target event rather than continuous monitoring.}
\label{tab:benchmark_comparison}
% \vspace{-3mm}
\end{table*}

As artificial intelligence integrates into continuous real-world environments like embodied robotics and wearable devices~\cite{cournan2016improving,ahn2023safefac,zhang2025language,lynch2023interactive,fang2025robix}, agents are expected to maintain always-on, hour-scale multimodal comprehension. However, traditional turn-based models~\cite{zhang2023video,lin2024video,maaz2024video,li2024llava} struggle in these dynamic settings. By design, they process pre-segmented video clips and passively wait for user prompts, which disrupts the causal continuity of real-world observations. Physical events rarely align with predefined temporal boundaries or user-initiated triggers. Consequently, turn-based agents often miss fleeting cues and fail to intervene autonomously when immediate action is required. Even low-latency systems like Doubao and GPT-Realtime-2~\cite{seed2026seed2,openai2026gptrealtime2} remain trapped in this passive prompt-response loop, lacking vigilance to self-initiate interactions. Genuine deployment therefore demands a paradigm shift toward interactive streaming architectures~\cite{lu2026aura,thinkingmachines2026interactionmodels,chen2024videollm,qian2025dispider}. These systems continuously ingest unbounded audio-visual streams, autonomously consolidate historical memory, and proactively determine when to speak, utilize tools, or remain silent without explicit human intervention.

Toward continuous, interactive, and long-horizon agentic streaming video understanding, we seek to establish an evaluation framework that exposes current limitations and guides future model development. However, existing evaluations have not kept pace with the expanding scope of streaming systems. Existing benchmarks provide valuable tests of long-video comprehension, online perception, or proactive response~\cite{fu2025video,lin2026streamingbench,wang2025omnimmi}, but generally assess these capabilities in isolation. Moreover, many streaming-video benchmarks use short clips and multiple-choice questions, making performance susceptible to language priors and recency shortcuts. Recent analysis shows that a baseline using only the last four frames can match substantially more complex streaming methods on such evaluations~\cite{shen2026simple}. High benchmark scores therefore do not necessarily establish sustained, causal understanding of an hour-scale stream. To close these evaluation gaps, we introduce \textit{StreamArena} and design its annotation protocol around the requirements of continuous deployment. First, 243 full-length videos with an average duration of 88.8 minutes expose models to genuinely long temporal horizons rather than isolated clips. Second, all questions require open-ended generation, removing answer-option cues that can mask failures in audio-visual grounding. Third, annotators assign separate timestamps to each query and every supporting evidence segment. This dual temporal grounding enforces causal access, measures the evidence-to-query gap, and specifies when a proactive response should occur. Questions within each video also preserve conversational continuity, enabling multi-turn evaluation over a shared stream. Quality control directly targets answerability and temporal correctness: two independent annotators answer each draft solely from the video and correct factual, linguistic, or timestamp errors, after which a third annotator conducts a blind audit. The pipeline removes approximately 27\% of the drafts and yields 3,646 validated tasks. These tasks measure four deployment capabilities: (1) real-time perception of unfolding events, (2) historical retrospection over distant evidence, (3) proactive interaction without a new user prompt, and (4) multimodal tool utilization grounded in the observed stream.

Across these four capabilities, existing systems reveal clear trade-offs. Turn-based MLLMs can reason over rich video evidence, but only after a query, and do not target proactive responses. AURA~\cite{lu2026aura} and MiniCPM-o~\cite{cui2026minicpm} keep only recent observations, enabling prompt responses but limiting memory; AURA drops from 25.4\% historical accuracy within five minutes to 10.5\% beyond thirty minutes. VST~\cite{guan2026video} replaces past visual evidence with a text summary, achieving 21.2\% on retrospection, but does not target Proactive or tool use. StreamForest~\cite{zeng2026streamforest} and ThinkStream~\cite{liu2026thinking} process video continuously, although their results indicate room for improvement: StreamForest obtains 14.4\% on retrospection, while ThinkStream scores 8.0\%, 7.5\%, 1.2\%, and 1.8\% across perception, retrospection, Proactive, and tool use. These results highlight the complementary strengths of existing approaches while showing that strong performance across all four capabilities remains challenging.

To address these limitations, we introduce \textit{StreamMind}, a simple yet effective two-tier architecture with independently scheduled, function-specific workers. A Front Worker dispatches requests, and Monitor Workers track future conditions without blocking interaction. Asynchronously, a Memory Writer stores hierarchical events, entity relations, and key frames, while Router, Recall, and Search Workers perform evidence-driven retrieval. This separation preserves visual evidence while moving memory construction and multistep reasoning off the response-critical path. StreamMind improves over the strongest streaming baseline for each capability by 58.4\% on real-time perception, 53.7\% on historical retrospection, 228.1\% on tool use, and 54.7\% on proactive interaction. With the same Qwen3.5-397B-A17B backbone, it reduces pooled query-to-answer latency by 66.2\%.

Our main contributions are summarized as follows:

\begin{itemize}
    \item We introduce \textit{StreamArena}, the first benchmark to jointly evaluate four capabilities over continuous audio-visual streams. It contains 243 full-length videos averaging 88.8 minutes and 3,646 manually validated, open-ended tasks with causal query and evidence timestamps.
    \item Extensive experiments across five classes of systems show that existing designs support perception, retrospection, Proactive, and tool use only partially or perform poorly when covering them jointly, revealing the conditions under which each design fails.
    \item We propose \textit{StreamMind}, a simple yet effective two-tier architecture that decouples interaction from memory and retrieval. It ranks first among streaming systems on all four capabilities, improving over the strongest corresponding baselines by 53.7\% to 228.1\% while reducing pooled response latency by 66.2\% under a shared backbone.
\end{itemize}

\section{Methodology}
\label{sec:methodology}

This section introduces StreamArena, an hour-scale causal benchmark, and StreamMind, an architecture decoupling responsive interaction from long-horizon memory.

\subsection{The StreamArena Benchmark}
\label{ssec:streamarena}

Unlike existing datasets that evaluate passive memory retention on offline clips, StreamArena provides a testbed for always-on streaming assistants. We outline its data curation protocol, statistical insights, and evaluation metrics below.

\subsubsection{Dataset Construction and Annotation}
\label{sssec:dataset_construction}

\paragraph{Video sourcing.}
We source YouTube videos from seven domains. Each video lasts at least 60 minutes, has a resolution of at least 1080p, and includes English or Chinese audio. We exclude sensitive, inappropriate, and political content.

\begin{figure*}[ht]
    \centering
    \includegraphics[width=0.95\linewidth]{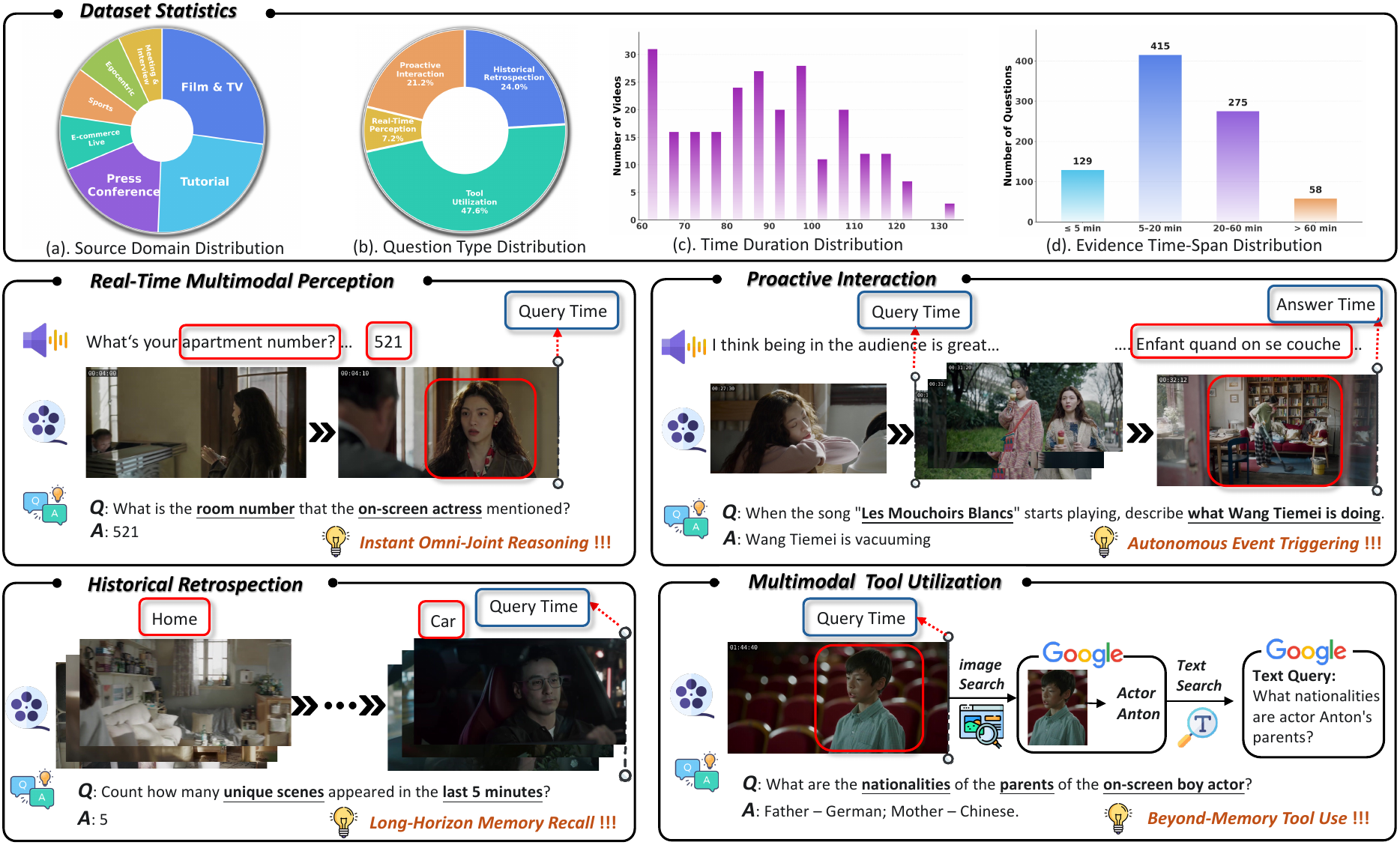}
    \caption{Overview of \textbf{StreamArena}. The top row summarizes (a) source domains, (b) task types, (c) video durations, and (d) temporal gaps between queries and supporting evidence. The remaining panels show one example of each evaluated capability. \evidencebox{}~\textcolor{evidencered}{\textbf{Red boxes}} mark the audio-visual evidence required to answer a question, while \timebox{}~\textcolor{timeblue}{\textbf{dark-blue boxes}} mark the query time and, for proactive tasks, the expected response time.}
    \label{fig:task_types}
% \vspace{-3mm}
\end{figure*}

\paragraph{Annotation protocol and question design.}
Thirty PhD-level annotators draft approximately 20 multi-turn question-answer pairs per video. Two independent cross-validators then answer and correct each draft, and a third independent annotator audits its question, answer, evidence, and timestamps. Answers are concise and objective, and every query and supporting evidence segment has an explicit timestamp. Questions from the same video preserve conversational continuity. Figure~\ref{fig:task_types} illustrates the four capabilities: (1) {real-time multimodal perception} requires joint reasoning over synchronized audio and video; (2) {historical retrospection} queries past events; (3) {multimodal tool utilization} requires Google Search for information unavailable from the stream or parametric memory; and (4) {proactive interaction} requires autonomous monitoring and alerting. This three-stage process retains approximately 73\% of drafts, yielding about 15 finalized pairs per video and 3,646 pairs in total.

\subsubsection{Dataset Statistics}
\label{sssec:dataset_statistics}

\paragraph{Scale and composition.}
Figure~\ref{fig:task_types}(a-d) summarizes StreamArena's 243 videos and 3,646 question-answer pairs across seven domains. Videos span 60 to 134.2 minutes (88.8 on average), with 263 perception, 877 retrospection, 1,732 tool-use, and 774 proactive tasks.

\paragraph{Temporal and evidence statistics.}
Historical-retrospection questions have a median evidence-to-query gap of 12.1 minutes, with an interquartile range of 5.7 to 25.7 minutes, and 49 questions place the nearest annotated evidence more than one hour before the query. Moreover, 189 questions (21.6\%) require joint attention to at least two distinct evidence segments, with the most demanding question spanning 15 segments. Table~\ref{tab:min_gap} aligns the temporal distribution with the strata used in Table~\ref{tab:main_results}: 20.6\% of HR questions fall beyond 30 minutes, while 12.0\% of proactive tasks require monitoring for more than 4 minutes.

\begin{table}[ht]
\centering
\footnotesize
\setlength{\tabcolsep}{3.6pt}
\renewcommand{\arraystretch}{1.02}
\begin{tabular}{@{}llrr@{}}
\toprule
\textbf{Capability} & \textbf{Layer} & \textbf{Horizon} & \textbf{\#Questions} \\
\midrule
HR   & L1 & $\leq 5$ min       & 181 (20.6\%) \\
     & L2 & 5 to 15 min        & 330 (37.6\%) \\
     & L3 & 15 to 30 min       & 185 (21.1\%) \\
     & L4 & $>30$ min          & 181 (20.6\%) \\
\midrule
Pro. & L1 & $\leq 30$ s        & 259 (33.5\%) \\
     & L2 & 30 s to 4 min      & 422 (54.5\%) \\
     & L3 & $>4$ min           &  93 (12.0\%) \\
\bottomrule
\end{tabular}
\caption{Temporal strata for historical retrospection (HR) and proactive interaction (Pro.).}
\label{tab:min_gap}
% \vspace{-4mm}
\end{table}

\subsubsection{Evaluation Metrics}
\label{sssec:evaluation_metrics}

Because StreamArena adopts open-ended question, exact-match accuracy is insufficient. Gemini~3.1~Pro serves as the LLM judge and makes a strict binary decision on whether each response contains the factual core of its  ground truth. Let $\mathcal{Q}$ represent the set of queries, with $N = |\mathcal{Q}|$. For a given query $q_i \in \mathcal{Q}$, let $\hat{a}_i$ be the agent's generated response and $a^*_i$ denote the ground truth. We formulate our evaluation across two primary dimensions.

\paragraph{Response accuracy and tool-enabled performance.}
For explicitly triggered tasks (e.g., historical retrospection, multimodal tool utilization), the primary objective is factual correctness. We define the accuracy ($\text{Acc}$) as:
\begin{equation}
\label{eq:acc}
    \text{Acc} = \frac{1}{N} \sum_{i=1}^N \mathbb{I}\Big(\text{Judge}(\hat{a}_i, a^*_i) = 1\Big),
\end{equation}
where the indicator function $\mathbb{I}(\cdot)$ yields $1$ if the LLM judge verifies that $\hat{a}_i$ strictly encapsulates the factual core of $a^*_i$, and $0$ otherwise. For tool-utilization tasks, this is a tool-enabled end-to-end answer metric; answer correctness alone does not establish whether or how a tool was invoked.

\paragraph{Temporal latency and triggering accuracy.}
For reactive tasks, we measure query-to-answer response latency as $L = t_{\text{resp}} - t_{\text{query}}$. This interval includes every operation triggered by the query, including routing, recall, external search, and model inference, while excluding continuous stream processing completed before the query arrives. For proactive tasks, the user first registers a monitoring instruction, after which the agent alerts autonomously when the target event occurs, without receiving another prompt at that moment. Let $t_{\text{pred}}$ be the video timestamp of the observation that triggers the alert and $t_{\text{gt}}$ the annotated event timestamp. We exclude subsequent response-generation latency from $t_{\text{pred}}$ so that the metric evaluates temporal vigilance rather than decoding speed. A proactive response is correct only if its content and trigger time are both valid:
\begin{equation}
\label{eq:proactive_acc}
\text{Proactive-Acc}=\frac{1}{N_{\mathrm{pro}}}\sum_{i=1}^{N_{\mathrm{pro}}}
\mathbb{I}\!\left[\text{Judge}(\hat a_i,a_i^*)=1\right]
\mathbb{I}\!\left[\operatorname{TimeOK}(t_i^{\mathrm{pred}},t_i^{\mathrm{gt}})=1\right],
\end{equation}
where $N_{\text{pro}}$ is the number of proactive tasks and $\operatorname{TimeOK}$ accepts $-0.5\,\text{s}\le t_i^{\mathrm{pred}}-t_i^{\mathrm{gt}}\le2.0\,\text{s}$. This formulation requires the agent to detect the event at the correct point in the stream regardless of its subsequent text generation speed.

\subsection{The StreamMind Architecture}
\label{ssec:streammind_architecture}

\begin{figure*}[t]
    \centering
    \includegraphics[width=0.95\linewidth]{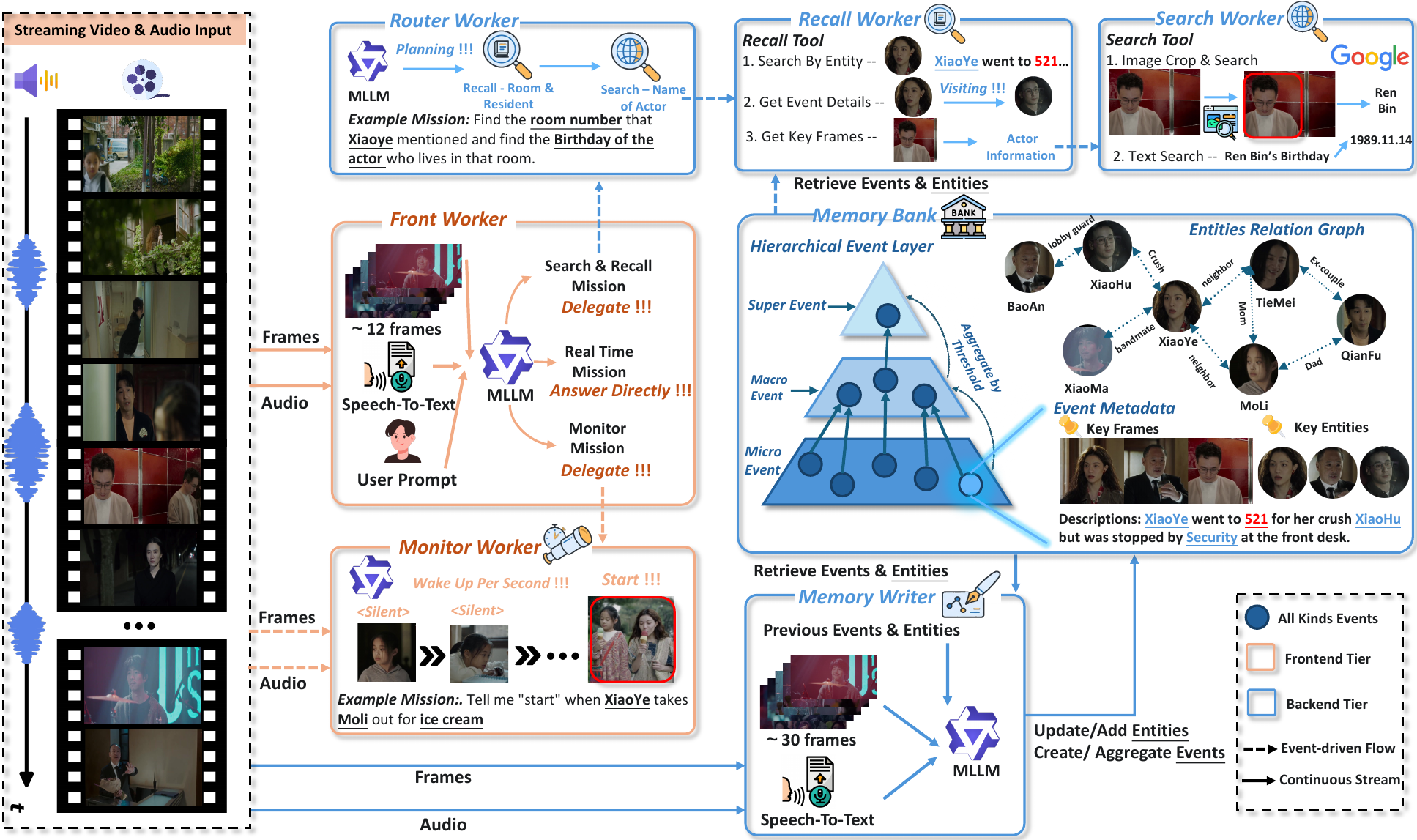}
    \caption{Overview of \textbf{StreamMind}. The \textcolor{orange!85!black}{\textbf{frontend}} handles interaction and monitoring; the \textcolor{backendblue}{\textbf{backend}} maintains memory and performs retrieval. The Memory Bank combines hierarchical events, entity relations, and key frames.}
    \label{fig:streammind}
    % \vspace{-4mm}
\end{figure*}

StreamMind decouples latency-sensitive interaction from long-horizon cognition through the two-tier architecture in Figure~\ref{fig:streammind}. The frontend interfaces with users and performs task dispatch, whereas the backend maintains persistent memory and executes retrieval-intensive reasoning. Function-specific workers communicate through on-demand requests while sharing memory updated only with observations available so far. This design lets the frontend respond immediately and access long-horizon evidence when needed.

\paragraph{Frontend interaction and dispatch.}
The Front Worker is the interaction gateway and high-level dispatcher. Given the user turn, recent causal observations, and conversation state, it selects among three paths. It answers immediately when context is sufficient, formulates a retrieval brief for the backend when the request requires historical or external evidence, or instantiates a Monitor Worker when the user asks the system to watch for a future condition. Each Monitor Worker maintains an independent lifecycle for its assigned condition and notifies the frontend only when that condition is satisfied. Consequently, persistent vigilance does not occupy the Front Worker or delay unrelated interactions.

\paragraph{Asynchronous memory construction.}
In parallel with frontend interaction, the Memory Writer continuously transforms incoming frames and speech observations into the Memory Bank shown in Figure~\ref{fig:streammind}. Its hierarchical event layer organizes local actions into micro events and progressively aggregates them into macro and super events. The complementary entity relation graph links recurring people, objects, locations, and their relations across time. Events retain key entities, temporal boundaries, textual descriptions, and representative frames, preserving both compact semantic structure and retrievable visual evidence. Memory construction is independent of user queries, so the stream is consolidated before a later question reveals which evidence will be needed.

\paragraph{Backend retrieval and coordination.}
When the Front Worker delegates a mission, the Router Worker becomes the coordinator. It decomposes the request into Recall and Search subtasks, executes independent subtasks concurrently, inspects their returned evidence, and either issues another round of retrieval or synthesizes the final response. The Recall Worker performs content-addressable retrieval over the event hierarchy and entity graph, returning relevant descriptions, relations, and historical key frames. The Search Worker grounds visible entities through image retrieval and acquires information unavailable from the stream through external text search. Their composition supports multistage requests: the Router can first recover an earlier event and its key frame through Recall, then use that frame as the visual anchor for Search. Recall and Search therefore remain specialized and need not communicate directly; the Router mediates their evidence exchange and global reasoning state.

\paragraph{Decoupled execution.}
The solid and dashed flows in Figure~\ref{fig:streammind} summarize two execution modes. Continuous ingestion drives monitoring and memory construction, while event-driven messages create Monitors or activate backend recall or external search only when required. Every query-driven memory read is bounded by the corresponding query time, preserving causal access to the stream. By pairing a non-thinking Front Worker with thinking-enabled backend workers, StreamMind supports low-latency interaction, proactive monitoring, historical recall, and external tool use.

% Auto-generated by tools/generate_main_results_table.py. DO NOT EDIT NUMBERS MANUALLY.
% Capability averages are weighted by the number of scored questions in each bucket.
\begin{table*}[t]
\centering
\footnotesize
\setlength{\tabcolsep}{2.5pt}
\renewcommand{\arraystretch}{1.15}
\begin{tabular}{@{}l l c c ccccc c cccc@{}}
\toprule
\multirow{2}{*}{\textbf{Method}} & \multirow{2}{*}{\textbf{Backbone}}
 & \multirow{2}{*}{\textbf{ASR}}
 & \multirow{2}{*}{\textbf{RTP} $\uparrow$}
 & \multicolumn{5}{c}{\textbf{HR} $\uparrow$}
 & \multirow{2}{*}{\textbf{Tool} $\uparrow$}
 & \multicolumn{4}{c}{\textbf{Pro.} $\uparrow$} \\
\cmidrule(lr){5-9}\cmidrule(lr){11-14}
 & & & & Avg. & L1 & L2 & L3 & L4
   & & Avg. & L1 & L2 & L3 \\
\midrule
\multirow{2}{*}{Human$^{\mathsection}$} & \multirow{2}{*}{--} & \multirow{2}{*}{--} & \multirow{2}{*}{91.8} & $80.7^{\circlearrowleft}$ & 84.2 & 82.1 & 79.5 & 75.8 & \multirow{2}{*}{95.2} & \multirow{2}{*}{91.5} & \multirow{2}{*}{94.6} & \multirow{2}{*}{91.3} & \multirow{2}{*}{83.4} \\
 & & & & $63.4^{\rightarrow}$ & 78.6 & 68.9 & 56.4 & 45.1 & & & & & \\
\midrule
\multicolumn{14}{l}{\textit{(A) Offline turn-based MLLMs}} \\
Qwen3.5-397B-A17B & -- & \cmarkT & 44.1 & 41.5 & 44.2 & 47.3 & 39.5 & 30.4 & 62.2 & -- & -- & -- & -- \\
MiMo-V2.5 & -- & \cmarkF & 38.0 & 35.8 & 29.3 & 45.5 & 31.9 & 28.7 & 47.9 & -- & -- & -- & -- \\
Kimi-K2.6 & -- & \cmarkT & 47.9 & 43.8 & 42.5 & 53.3 & 38.4 & 33.1 & 60.9 & -- & -- & -- & -- \\
Gemini 3.5 Flash$^{\dagger}$ & -- & \cmarkF & 51.3 & 51.4 & 45.8 & 57.8 & 51.4 & 45.3 & 70.8 & -- & -- & -- & -- \\
Qwen3.5-Omni$^{\dagger}$ & -- & \cmarkF & 41.8 & 35.8 & 31.5 & 44.8 & 34.1 & 25.4 & 49.4 & -- & -- & -- & -- \\
\midrule
\multicolumn{14}{l}{\textit{(B) Recent-window methods}} \\
AURA$^{\ddagger}$ & Qwen3-VL-8B-Instruct & \cmarkT & 28.1 & 22.7 & 25.4 & 27.0 & 24.3 & 10.5 & -- & 3.7 & 8.9 & 1.4 & 0.0 \\
MiniCPM-o-4.5$^{\ddagger}$ & Qwen3-8B & \cmarkF & 22.1 & 9.8 & 10.5 & 13.3 & 7.6 & 5.0 & 17.1 & 7.5 & 7.3 & 8.1 & 5.4 \\
\midrule
\multicolumn{14}{l}{\textit{(C) Text-summary methods}} \\
VST$^{\ddagger}$ & Qwen2.5-VL-Instruct & \cmarkT & 24.0 & 21.2 & 22.1 & 24.8 & 23.2 & 11.6 & -- & -- & -- & -- & -- \\
\midrule
\multicolumn{14}{l}{\textit{(D) Model-internal compression methods}} \\
StreamForest & Qwen2-7B & \cmarkT & 17.9 & 14.4 & 19.3 & 13.6 & 14.1 & 11.0 & -- & -- & -- & -- & -- \\
ThinkStream & Qwen2.5-
VL-3B & \cmarkT & 8.0 & 7.5 & 12.2 & 7.9 & 5.4 & 4.4 & 1.8 & 1.2 & 2.7 & 0.5 & 0.0 \\
\midrule
\multicolumn{14}{l}{\textit{(E) Ours}} \\
\rowcolor{gray!12}
\textbf{StreamMind (Ours)} & \textbf{Qwen3.5-397B-A17B} & \cmarkT & \textbf{44.5} & \textbf{34.9} & \textbf{31.5} & \textbf{46.7} & \textbf{34.6} & \textbf{17.1} & \textbf{56.1} & \textbf{11.6} & \textbf{16.6} & \textbf{9.5} & \textbf{7.5} \\
\bottomrule
\end{tabular}
\caption{Accuracy (\%) on \textbf{StreamArena}. HR levels denote evidence-to-query gaps of $\leq5$, 5 to 15, 15 to 30, and $>30$ minutes; Pro. levels denote monitoring horizons of $\leq30$ seconds, 30 seconds to 4 minutes, and $>4$ minutes. Avg. is question-weighted. ASR indicates explicit transcript input. For Human$^{\mathsection}$, $^{\circlearrowleft}$ and $^{\rightarrow}$ denote HR with and without rewatching, corresponding to offline and streaming conditions, respectively. $^{\dagger}$ denotes a closed-source model, $^{\ddagger}$ an author-finetuned backbone, and -- not applicable or an unsupported capability.}
\label{tab:main_results}
% \vspace{-4mm}
\end{table*}

\section{Experiments}
\label{sec:experimental}

\subsection{Experimental Setups}
\label{ssec:experimental_setups}

We evaluate five architecture groups on StreamArena: (A) offline turn-based MLLMs, including Qwen3.5-397B-A17B, MiMo-V2.5, Kimi-K2.6, Gemini~3.5~Flash, and Qwen3.5-Omni; (B) recent-window methods, namely AURA~\cite{lu2026aura} and MiniCPM-o-4.5~\cite{cui2026minicpm}; (C) the text-summary method VST~\cite{guan2026video}; (D) model-internal compression methods, namely StreamForest~\cite{zeng2026streamforest} and ThinkStream~\cite{liu2026thinking}; and (E) our StreamMind. Gemini~3.1~Pro scores every supported output using the same strict factual-core decision, and systems with proactive support share the same timing rule. To reflect each system's design, we retain its native inference mechanism rather than forcing all methods into continuous ingestion. Although several baselines are architecturally designed for streaming input, their released evaluation interfaces reconstruct a method-specific causal prefix or recent window when a query or monitoring task arrives and preserve no hidden video state across tasks. We use these interfaces for capability comparison while distinguishing query-triggered evaluation from continuous, stateful ingestion. StreamMind ingests frames sequentially from $t=0$ at 2 fps while memory construction and monitoring continue between user turns.

\paragraph{Evaluation protocols.}
Offline models uniformly sample up to 128 frames from each causal prefix. AURA and MiniCPM-o use the latest 30 seconds, while VST summarizes at most 384 causal-prefix frames. StreamForest and ThinkStream reconstruct causal prefixes using up to 2,048 frames and 120 two-frame chunks, respectively. StreamMind continuously ingests at 2 fps and preserves state. Reactive inputs never contain frames after the query time, and dialogue history is preserved within each video.

\begin{figure*}[t]
    \centering
    \includegraphics[width=\linewidth]{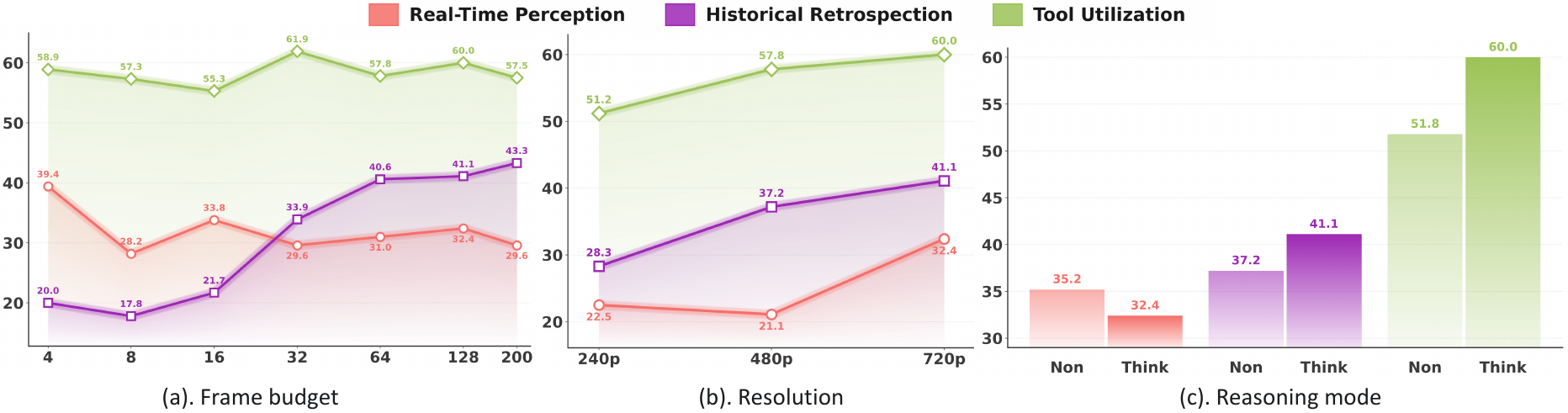}
    \caption{Accuracy (\%) on the diagnostic subset under changes to (a) frame count, (b) resolution, and (c) reasoning mode.}
    \label{fig:diagnostic_ablation}
% \vspace{-3mm}
\end{figure*}

\subsection{Main Results}
\label{ssec:main_results}

Table~\ref{tab:main_results} compares five architecture groups across the four target capabilities and four HR horizons. StreamMind ranks first among streaming systems in every capability, while the temporal breakdown reveals how alternative memory designs degrade as evidence becomes more distant.

\paragraph{Unified capability coverage.}
The human reference reaches 91.8\% on RTP, 95.2\% on Tool, and 91.5\% on Proactive. On HR, human accuracy decreases from 80.7\% with rewatching to 63.4\% without it, illustrating the difficulty of retaining hour-scale evidence under streaming access. Offline turn-based MLLMs provide strong reactive references, but they do not support proactive interaction and do not maintain state as the video unfolds. Other systems cover only part of the target capability space: recent windows favor recent perception, text summaries sacrifice visual evidence, and model-internal compression remains weak on open-ended long-horizon questions. StreamMind is the only evaluated system that combines continuous ingestion with support for all four capabilities. Relative to the strongest streaming baseline that supports each capability, it improves RTP by 58.4\%, HR by 53.7\%, Tool by 228.1\%, and Proactive by 54.7\% relatively. These results show that persistent memory, proactive monitoring, and retrieval are complementary components rather than interchangeable forms of temporal context.

\paragraph{Long-horizon comprehension.}
StreamMind improves every HR gap bucket over the strongest streaming baseline for that bucket, with relative gains ranging from 24.0\% to 73.0\%. Recent-window methods lose observations outside their recent context, whereas text-summary and model-internal compression methods retain broader temporal coverage at the cost of fine-grained evidence. StreamMind instead retrieves both structured event descriptions and associated key frames from persistent memory. Its advantage across all four gap ranges indicates that long-horizon streaming benefits from preserving retrievable multimodal evidence rather than relying on a single transient context or textual summary.

\paragraph{Tool use and proactive interaction.}
Tool questions require the system to ground visual referent, identify information gap, invoke an appropriate retrieval operation, and integrate external evidence. StreamMind's relative improvement of 228.1\% over the strongest streaming baseline on Tool reflects the value of explicit coordination among Router, Recall, and Search Workers. Proactive interaction presents a different challenge because the system receives no new prompt when the target event occurs. StreamMind obtains a relative improvement of 54.7\% through independently scheduled Monitor Workers, showing that persistent vigilance and reactive question answering require distinct execution paths.

% Auto-generated by tools/generate_latency_table.py. DO NOT EDIT NUMBERS MANUALLY.
% Mean wall-clock end-to-end latency (seconds) per question type.
\begin{table}[t]
\centering
\footnotesize
\setlength{\tabcolsep}{3.5pt}
\renewcommand{\arraystretch}{1.15}
\begin{tabular}{@{}l c ccc@{}}
\toprule
\multirow{2}{*}{\textbf{Method}} & \multirow{2}{*}{\textbf{ACC} $\uparrow$} & \multicolumn{3}{c}{\textbf{Mean latency (s)} $\downarrow$} \\
\cmidrule(lr){3-5}
 & & A (\textsc{RTP}) & B (\textsc{HR}) & D (\textsc{Tool}) \\
\midrule
Gemini 3.5 Flash$^{\dagger}$ & 63.1 & 86.9 & 123.9 & 139.5 \\
Kimi-K2.6 & 54.5 & 162.8 & 187.9 & 128.2 \\
Qwen3.5-397B-A17B & 54.2 & 83.8 & 117.4 & 62.8 \\
Qwen3.5-Omni$^{\dagger}$ & 44.6 & 50.8 & 56.4 & 103.7 \\
MiMo-V2.5 & 43.3 & 224.5 & 266.5 & 187.4 \\
\rowcolor{gray!12}
\textbf{StreamMind (Ours)} & \textbf{48.6} & \textbf{12.9} & \textbf{30.7} & \textbf{28.1} \\
\bottomrule
\end{tabular}
\caption{Mean query-to-answer latency in seconds and pooled accuracy (\%) on RTP, HR, and Tool questions. $^{\dagger}$ denotes a closed-source model.}
\label{tab:latency_comparison}
% \vspace{-3mm}
\end{table}

\subsection{Online Response Latency}
\label{ssec:online_latency}

Table~\ref{tab:latency_comparison} reports the wall-clock interval from query arrival to final-answer generation. The most controlled comparison uses Qwen3.5-397B-A17B as the shared backbone. Relative to query-triggered offline inference, StreamMind reduces response latency by 84.6\% on RTP, 73.9\% on HR, and 55.3\% on Tool. Weighting the three categories by their numbers of questions gives a 66.2\% relative reduction, from 81.4 to 27.5 seconds, while StreamMind retains 89.7\% of the pooled accuracy. Comparisons with other backbones provide practical system references, but their latency also reflects differences in model scale and serving infrastructure. The reduction arises from state reuse rather than free computation. Offline models sample and encode a visual context independently for every question. StreamMind continuously constructs memory before a query arrives, then retrieves task-relevant events and key frames on demand. It therefore moves persistent perception and memory construction away from the response-critical path while including all query-triggered routing, recall, search, and inference in the reported latency.

\subsection{Benchmark Diagnostic Analysis}
\label{ssec:benchmark_diagnostics}

We further audit common input shortcuts on a 616-question subset of StreamArena, comprising 71 real-time perception, 180 historical retrospection, and 365 tool-utilization questions. All diagnostics use Qwen3.5-397B-A17B with the same scoring protocol. Unless varied, the model receives 128 frames at 720p with ASR and uses thinking mode.

\begin{table}[t]
\centering
\footnotesize
\setlength{\tabcolsep}{4.2pt}
\renewcommand{\arraystretch}{1.12}
\begin{tabular}{@{}lrrrr@{}}
\toprule
\textbf{Available input} & \textbf{RTP} & \textbf{HR} & \textbf{Tool} & \textbf{Overall} \\
\midrule
ASR only                  &  4.2 & 16.7 & 42.5 & 30.5 \\
Visual only (128 frames)  & 26.8 & 37.8 & \textbf{60.5} & 50.0 \\
Visual + ASR (128 frames) & \textbf{32.4} & \textbf{41.1} & 60.0 & \textbf{51.3} \\
\bottomrule
\end{tabular}
\caption{Accuracy (\%) by input modality on the diagnostic subset. Visual conditions use 128 frames.}
\label{tab:subset_modality}
% \vspace{-4mm}
\end{table}

\paragraph{Modality availability.}
Table~\ref{tab:subset_modality} highlights the multimodal nature of RTP questions. ASR-only input achieves 4.2\% accuracy, while visual-only input reaches 26.8\%. Combining visual input with ASR further improves accuracy to 32.4\%, a gain of 5.6 percentage points over visual input alone. This result shows that visual evidence is essential for real-time perception, while speech provides complementary information that cannot be recovered from sampled frames alone.

\paragraph{Evidence budgets and reasoning mode.}
Figure~\ref{fig:diagnostic_ablation} examines temporal coverage, spatial fidelity, and reasoning mode. Increasing the frame budget primarily benefits HR, while higher resolution improves all three tasks. Thinking further improves HR and Tool but slightly reduces RTP, suggesting that extended reasoning aids evidence integration at the cost of immediate perceptual precision. These trends support StreamMind's role-specific reasoning configuration.

\section{Discussion and Future Work}
\label{sec:discussion}

\paragraph{Long-horizon memory requires future-aware retention.}
StreamMind improves HR across temporal strata, but its 63.4\% decrease from L2 to L4 shows that long-horizon memory is not a capacity problem. Since future queries are unknown during ingestion, future work should learn from retrieval outcomes to estimate evidence utility and preserve uncertain multimodal details at adaptive fidelity.

\paragraph{Backbone scale and comparison scope.}
StreamMind uses Qwen3.5-397B-A17B because its workers require reliable instruction following for structured memory construction, tool routing, and condition monitoring. Since several streaming baselines use smaller backbones, their comparison with StreamMind reflects both model capacity and system design. We therefore do not attribute every accuracy gain to the architecture alone. Instead, the same-backbone offline Qwen3.5-397B-A17B comparison isolates the system-level trade-off: StreamMind retains 89.7\% of pooled accuracy while reducing response latency by 66.2\% through persistent state reuse. Proactive interaction has no direct offline counterpart because it requires continuous monitoring without a new user query. Evaluating smaller backbones and individual architectural components remains important future work.

\section{Conclusion}
\label{sec:conclusion}

We introduce StreamArena, an hour-scale benchmark that jointly evaluates real-time multimodal perception, historical retrospection, multimodal tool utilization, and proactive interaction under causal access. Its open-ended questions, timestamped evidence, and continuous protocol expose shortcuts that remain hidden in short-clip or multiple-choice evaluation. Experiments reveal a central tension between responsive interaction and persistent multimodal understanding. StreamMind addresses this tension through a decoupled architecture that combines frontend interaction with asynchronous memory construction, historical recall, and external search. It improves all four capabilities over existing streaming baselines and reduces query-to-answer latency by reusing state accumulated before each query. The remaining gaps in hour-scale recall, proactive monitoring, and continuous processing efficiency identify concrete directions toward practical always-on multimodal agents.

\newpage
\bibliographystyle{unsrt}
\bibliography{ref}

\newpage
\appendix

\section{Related work}
\label{sec:related-work}

\paragraph{Interactive streaming video understanding.}
Traditional video language models target offline, turn-based question answering over pre-segmented clips~\cite{zhang2023video,lin2024video,maaz2024video}. To extend comprehension to longer recordings, MemDreamer~\cite{chen2026memdreamer}, M3-Agent~\cite{long2025seeing}, EgoRAG~\cite{yang2025egolife}, and EGAgent~\cite{rege2026agentic} organize visual evidence into hierarchical summaries, episodic and semantic memories, or entity-centric graphs for subsequent retrieval. These methods primarily study memory representation and retrieval after video processing, with less attention to interleaving memory construction and user queries under strict causal and latency constraints. Interactive streaming instead requires a query at time $t$ to access only observations up to $t$. Streaming architectures support this causal, unbounded setting through hierarchical memory or selective token retention~\cite{zeng2026streamforest}, but passive compression alone cannot determine when to speak, interrupt, or remain silent. Recent frameworks add triggering modules or dual-model systems for proactive interaction~\cite{thinkingmachines2026interactionmodels,lu2026aura}, while StreamMind further decouples asynchronous memory construction from the real-time interaction path.

\paragraph{Video understanding benchmarks.}
Video-MME~\cite{fu2025video}, LongVideoBench~\cite{wu2024longvideobench}, and MLVU~\cite{zhou2025mlvu} test offline comprehension of long videos but omit causal, real-time constraints. StreamingBench~\cite{lin2026streamingbench}, OVO-Bench~\cite{niu2025ovo}, OVBench~\cite{huang2025online}, RTV-Bench~\cite{xun2026rtv}, and OST-Bench~\cite{lin2026ost} study online understanding, yet short clips and multiple-choice questions introduce recency bias and language priors. Minimal baselines can consequently perform well without long-horizon reasoning~\cite{shen2026simple}. OmniMMI~\cite{wang2025omnimmi} and QIVD (Qualcomm)~\cite{pourreza2025can} assess real-time multimodal interaction, but their short videos cannot measure sustained operation. StreamArena combines hour-scale videos, open-ended human-validated questions, proactive behavior, and external tool use.

\section{StreamArena details}
\label{app:streamarena_details}

\subsection{Data collection and annotation pipeline}
\label{app:annotation_pipeline}

We use a three-stage annotation pipeline with 30 PhD-level annotators.

\paragraph{Video sourcing and filtering.}
Annotators collect YouTube videos from seven domains. Each video lasts at least 60 minutes and has a resolution of at least 1080p. Screening excludes sensitive, inappropriate, and political content.

\paragraph{Initial generation.}
Annotators draft approximately 20 question-answer pairs per video. Each task contains the question, query timestamp, reference answer, and timestamped audio-visual evidence. The two quality-control stages retain approximately 73\% of drafts, producing about 15 finalized pairs per video and 3,646 pairs over 243 videos.

\paragraph{First-round cross-validation.}
Two annotators who do not participate in drafting independently answer each question from the video. They correct factual errors, ambiguous wording, and temporal misalignment.

\paragraph{Second-round independent audit.}
A third annotator, independent of drafting and cross-validation, audits each revised question, answer, and timestamp for consistency with the annotation guidelines. The final retention rate is approximately 73\%.

\subsection{Task taxonomy}
\label{app:task_taxonomy}

Every question in StreamArena belongs to one of four capabilities, each targeting a distinct axis of streaming competence. Below we give the operational definition, canonical query template, evidence structure, and scoring rule of every capability. Throughout this subsection we denote the video stream by $\mathcal{S} = \{s_t\}_{t=0}^{T}$, where $s_t$ is the audio-visual frame available at time $t$, and write $\mathcal{S}_{a:b} = \{s_t\}_{t=a}^{b}$ for the causal prefix or window of interest. A question is a tuple $(q_i, t^{q}_i, a^*_i, \mathcal{E}_i)$ that specifies the query text $q_i$, its query time $t^{q}_i$, the ground-truth answer $a^*_i$, and the annotated evidence set $\mathcal{E}_i$. The agent implements a policy $f$ that emits a response $\hat{a}_i$; concrete signatures for $f$ are given below. Every capability reuses the LLM-judge accuracy of Eq.~\eqref{eq:acc} unless noted otherwise, and concrete examples are shown in Figure~\ref{fig:task_types}.

\paragraph{Real-time multimodal perception (RTP).}
RTP questions probe the ability of the agent to answer a query using audio-visual evidence in a short window around the query time. A typical question refers to an on-screen object, a person's action, an ambient sound, or a spoken utterance whose grounding cue is visible or audible within a few seconds of $t^{q}_i$. Formally, the evidence set is a short window
\begin{equation}
\label{eq:rtp_evidence}
    \mathcal{E}_i^{\text{RTP}} \subset \mathcal{S}_{t^{q}_i - \delta \,:\, t^{q}_i + \delta}, \quad \delta \le 10 \text{ s},
\end{equation}
and the agent produces a response
\begin{equation}
\label{eq:rtp_policy}
    \hat{a}_i = f_{\text{RTP}}\big(q_i,\, \mathcal{S}_{0:t^{q}_i}\big).
\end{equation}
Scoring uses the accuracy in Eq.~\eqref{eq:acc}. This capability isolates \textit{joint audio-visual grounding over a short causal context}: an agent that ignores the audio track, undersamples the stream, or fails to preserve transient cross-modal cues cannot recover the required evidence.

\paragraph{Historical retrospection (HR).}
HR questions require the agent to recall factual information from earlier in the stream. Each question is grounded in one or more timestamped supporting evidence segments
\begin{equation}
\label{eq:hr_evidence}
    \mathcal{E}_i^{\text{HR}} = \{(t^{e}_{i,k}, \psi_{i,k})\}_{k=1}^{K_i},
    \qquad t^{e}_{i,k} < t^{q}_i,
\end{equation}
where $\psi_{i,k}$ is an annotation that briefly describes the supporting cue at time $t^{e}_{i,k}$ and $K_i \ge 1$. These descriptions are used only to identify and audit the supporting evidence and are not provided to the evaluated agent. The evidence-to-query gap
\begin{equation}
\label{eq:hr_gap}
    \Delta^{\text{HR}}_i = t^{q}_i - \max_{k} t^{e}_{i,k}
\end{equation}
can exceed one hour (Table~\ref{tab:min_gap}), and the agent produces $\hat{a}_i = f_{\text{HR}}(q_i, \mathcal{S}_{0:t^{q}_i})$ scored by Eq.~\eqref{eq:acc}. HR is designed to stress \textit{long-horizon memory}: the agent needs to persist observations, index them by content, and retrieve them under an open-ended query.

We further label every HR question with a \textit{reasoning pattern} $r_i$ that specifies how the agent combines the $K_i$ evidence segments:
\begin{enumerate}[label=(\roman*), leftmargin=1.8em, itemsep=2pt, topsep=2pt]
    \item \textbf{Single point} ($K_i = 1$): a single evidence segment is sufficient (e.g., ``what room number did the actress mention?'').
    \item \textbf{Multi-point count} ($a^*_i \in \mathbb{N}$): the agent counts the number of times an event or object occurs across multiple segments (e.g., ``how many unique scenes appeared in the last five minutes?'').
    \item \textbf{Multi-point recall} ($a^*_i = \{\psi_{i,k}\}_{k}$): multiple evidence segments each contribute a fact that jointly forms the answer.
    \item \textbf{Multi-point compare}: the agent compares quantities or attributes across two or more evidence segments (e.g., which of two prices was higher).
    \item \textbf{Temporal range}: the answer summarizes a continuous interval $[t^{e}_{i,1}, t^{e}_{i,K_i}] \subset [0, t^{q}_i]$ rather than a set of isolated points.
\end{enumerate}

\paragraph{Multimodal tool utilization (Tool).}
Tool questions ask about entities, facts, or attributes that are not observable from the video stream itself. The evaluation protocol therefore equips the agent with external image and text search, represented by an operator $\pi: \mathcal{X} \rightarrow \mathcal{Y}$; the agent's response follows
\begin{equation}
\label{eq:tool_policy}
    \hat{a}_i = f_{\text{Tool}}\big(q_i,\, \mathcal{S}_{0:t^{q}_i},\, \{\pi(x_j)\}_{j}\big),
\end{equation}
where $\{x_j\}_j$ are search queries selected by the agent and $\{\pi(x_j)\}_j$ are the returned external observations. As illustrated in Figure~\ref{fig:task_types}, the canonical workflow first grounds a visual referent (e.g., a person on screen at $t^{q}_i$), then applies image search to obtain a textual handle (e.g., the actor's name), and finally issues a text query to retrieve the requested attribute. Scoring uses Eq.~\eqref{eq:acc}. We therefore report Tool accuracy as tool-enabled end-to-end answer accuracy, rather than treating a correct final answer alone as proof that a particular tool was invoked.

\paragraph{Proactive interaction (Pro).}
The user issues a monitoring instruction $q_i$ at time $t_i^q$, before a target event annotated at time $t_i^{\text{gt}}$. No additional prompt is provided when the event occurs. After registering the instruction, the agent continuously monitors the subsequent stream and autonomously emits an alert; formally,
\begin{equation}
\label{eq:pro_policy}
    \{(t, o_{i,t})\}_{t \ge t_i^q} = f_{\text{Pro}}(q_i,\mathcal{S}_{0:t}),
    \qquad o_{i,t} \in \{\varnothing\} \cup \mathcal{V}_{\mathrm{alert}},
\end{equation}
where $o_{i,t}=\varnothing$ denotes no alert and $o_{i,t}\in\mathcal{V}_{\mathrm{alert}}$ an autonomous notification associated with the registered task. The immediate acknowledgment of $q_i$, if any, is not an alert and is excluded from this output stream. We define $t_i^{\mathrm{pred}}=\min\{t\ge t_i^q:o_{i,t}\ne\varnothing\}$, with $t_i^{\mathrm{pred}}=+\infty$ if no alert is produced. Because event annotations and sampled observations have finite temporal granularity, the timing rule allows a small lead while imposing a strict delay bound:
\begin{equation}
\label{eq:pro_time_window}
    \operatorname{TimeOK}(t_i^{\mathrm{pred}},t_i^{\mathrm{gt}})
    =\mathbb{I}\big(-0.5\,\text{s}\le t_i^{\mathrm{pred}}-t_i^{\mathrm{gt}}\le2.0\,\text{s}\big).
\end{equation}
The metric is Proactive-Acc in Eq.~\eqref{eq:proactive_acc}. This capability isolates \textit{always-on temporal vigilance}: an agent that only reasons when explicitly prompted cannot receive any credit.

\subsection{Complete dataset statistics}
\label{app:statistics_full}

\paragraph{Video-level metadata.}
All 243 videos have a duration of $\ge 60$ minutes, with a mean of 88.8 minutes and a maximum of 134.2 minutes. The audio-visual content is dominated by Mandarin: 189 videos carry a Chinese audio track, 25 are in English, 26 mix Chinese and English, and 3 are in another language. 17 of the 243 videos are recorded live streams, while the remaining 226 are non-live long-form recordings. Per-domain video counts and mean durations are reported in Table~\ref{tab:domain_capability}, and the per-domain distribution of the finer L2 sub-categories is reported in Table~\ref{tab:l2_subdomain}.

\paragraph{Per-domain composition.}
Table~\ref{tab:domain_capability} lists, for each of the seven source domains, the number of videos, the mean video duration, and the number of question-answer pairs of each capability. The four capabilities are abbreviated as \textit{RTP} (real-time multimodal perception), \textit{HR} (historical retrospection), \textit{Tool} (multimodal tool utilization), and \textit{Pro.} (proactive interaction). Film \& TV and Tutorial together account for roughly half of the videos, while every remaining domain contributes at least 17 videos and 262 questions, providing sufficient statistical mass for per-domain analysis. Table~\ref{tab:l2_subdomain} refines this view by reporting the L2 sub-categories inside each domain.

\begin{table*}[ht]
\centering
\small
\setlength{\tabcolsep}{6pt}
\begin{tabular}{lrrrrrrr}
\toprule
Domain & \#Videos & Mean Dur. (min) & RTP & HR & Tool & Pro. & Total Q \\
\midrule
Film \& TV            & 66 & 90.0 & 66 & 239 & 438 & 207 & 950 \\
Tutorial              & 57 & 86.4 & 77 & 210 & 413 & 201 & 901 \\
Press Conference      & 43 & 82.5 & 36 & 153 & 312 & 123 & 624 \\
E-commerce Live       & 21 & 91.8 & 20 &  85 & 128 &  66 & 299 \\
Egocentric            & 20 & 87.7 & 22 &  57 & 146 &  49 & 274 \\
Sports                & 19 & 97.8 & 26 &  67 & 171 &  72 & 336 \\
Meeting \& Interview  & 17 & 96.0 & 16 &  66 & 124 &  56 & 262 \\
\midrule
Total                 & 243 & 88.8 & 263 & 877 & 1{,}732 & 774 & 3{,}646 \\
\bottomrule
\end{tabular}
\caption{Per-domain video count, mean duration, and per-capability question counts on StreamArena.}
\label{tab:domain_capability}
\end{table*}

\begin{table*}[ht]
\centering
\small
\setlength{\tabcolsep}{6pt}
\begin{tabularx}{\linewidth}{lX}
\toprule
Domain & Sub-category (\#videos) \\
\midrule
Film/TV            & Movie (40); Anime (16); TV series (10) \\
Tutorial              & Programming / software tutorial (47); K-12 or academic subjects (6); Arts and crafts (3); Game walkthrough (1) \\
Press Conference      & Film or game launch (22); Consumer tech launch (16); Automotive launch (5) \\
E-commerce Live       & General livestream selling (19); Food and groceries (1); Cosmetics (1) \\
Egocentric            & Travel vlog (16); Daily-life vlog (2); Retail or office surveillance (1); Game streaming (1) \\
Sports                & Basketball (15); Other ball sports (4) \\
Meeting/Interview  & Academic lecture (6); Celebrity interview or podcast (5); Live news broadcast (5); Politics or economics forum (1) \\
\bottomrule
\end{tabularx}
\caption{L2 sub-category distribution inside each StreamArena source domain. Numbers in parentheses report the video count of the sub-category.}
\label{tab:l2_subdomain}
\end{table*}

\paragraph{Historical-retrospection evidence complexity.}
For every HR question, annotators mark the video segments that supply the answer and tag the reasoning pattern (defined in Section~\ref{app:task_taxonomy}). Table~\ref{tab:reasoning_pattern} reports the pattern distribution over all 877 HR questions, and Table~\ref{tab:n_segments} reports the number of distinct evidence segments per question. The two views are complementary but not identical: while 312 questions ($877-565$) follow a multi-point or temporal-range reasoning pattern, 189 (21.6\%) require the agent to jointly attend to at least two distinct evidence segments. Some multi-point-count and temporal-range questions can be resolved from a single internally structured segment, such as a continuous statistics panel or a time-lapse shot. The hardest examples span up to 15 segments, ruling out any purely single-shot solution.

\begin{table}[ht]
\centering
\small
\begin{tabular}{lr}
\toprule
Reasoning pattern & \#Questions \\
\midrule
Single point                    & 565 \\
Multi-point count               & 174 \\
Multi-point recall              &  51 \\
Multi-point compare             &  14 \\
Temporal range                  &  73 \\
\bottomrule
\end{tabular}
\caption{Reasoning-pattern distribution over all 877 historical-retrospection questions.}
\label{tab:reasoning_pattern}
\end{table}

\begin{table}[ht]
\centering
\small
\begin{tabular}{lrrrrrrr}
\toprule
\#Segments & 1 & 2 & 3 & 4 & 5 & 6 & $\ge 7$ \\
\midrule
\#Questions & 688 & 99 & 34 & 25 & 7 & 10 & 14 \\
\bottomrule
\end{tabular}
\caption{Number of distinct video segments a historical-retrospection question requires the model to jointly attend to.}
\label{tab:n_segments}
\end{table}

\section{Detailed experimental setups}
\label{app:experimental_setups}

To ensure reproducibility and a fair cross-paradigm comparison, we document the evaluation protocol for every system in Table~\ref{tab:main_results}. Appendix~\ref{app:setup_taxonomy} distinguishes the evaluation harness from model design. Appendix~\ref{app:setup_offline} details Group~(A), Appendix~\ref{app:setup_pseudo_streaming} details Groups~(B) and~(C), and Appendix~\ref{app:setup_native_streaming} details Group~(D). Appendix~\ref{app:setup_streammind} presents the full configuration of StreamMind, and Appendix~\ref{app:setup_infrastructure} records computing infrastructure and reproducibility notes.

\subsection{Evaluation harness and model design}
\label{app:setup_taxonomy}

A key axis of StreamArena is whether a system truly ingests the video as an unbounded causal stream or merely simulates streaming through query-triggered replay. We formalize this distinction along two orthogonal dimensions.

\paragraph{Harness dimension.}
Given a query at time $t^{q}_i$, let $\mathcal{V}_i \subseteq \mathcal{S}$ denote the subset of frames that the harness makes available to the model to answer the $i$-th question. Under query-triggered replay, each method reconstructs its prescribed causal input on demand:
\begin{equation}
\label{eq:harness_replay}
    \mathcal{V}^{\text{replay}}_i
    =\operatorname{Sample}_i\!\left(\mathcal{S}_{a_i:t_i^q}\right),
    \qquad 0\le a_i\le t_i^q,
\end{equation}
where $a_i=0$ for prefix-replay methods and $a_i>0$ for recent-window methods. Under continuous ingestion,
\begin{equation}
\label{eq:harness_continuous}
    \mathcal{V}^{\text{stream}}_i =
    \left\{s_{k/f_{\mathrm{cap}}}:0\le k/f_{\mathrm{cap}}\le t_i^q\right\},
\end{equation}
where frames arrive sequentially at a fixed capture rate $f_{\text{cap}}$ from $t = 0$ and are irrevocably consumed by the model, which persists information in an internal state $h_t$ evolved by a causal recurrence
\begin{equation}
\label{eq:state_recurrence}
    h_t = \phi\!\left(h_{t - 1 / f_{\text{cap}}},\, s_t\right), \qquad h_0 = h_{\text{init}}.
\end{equation}
Replay can revisit and reconstruct its designated interval independently for every question. Continuous ingestion instead consumes each sampled observation once and preserves relevant information in state before its future utility is known.

\paragraph{Model dimension.}
Independent of the harness, a model applies one of two attention patterns to its visual input. Writing $z^{(v)}_t$ for the visual tokens at frame $t$ and $\text{Attn}(\cdot)$ for a masked attention operator, non-causal processing computes token representations
\begin{equation}
\label{eq:attn_noncausal}
    z^{(v)}_t \leftarrow \text{Attn}\!\left(z^{(v)}_t,\; \{z^{(v)}_{t'}\}_{t' \in [0, T_{\max}]}\right),
\end{equation}
whereas causal processing enforces the prefix-only mask
\begin{equation}
\label{eq:attn_causal}
    z^{(v)}_t \leftarrow \text{Attn}\!\left(z^{(v)}_t,\; \{z^{(v)}_{t'}\}_{t' \le t}\right).
\end{equation}

Table~\ref{tab:protocol_taxonomy} classifies every system in Table~\ref{tab:main_results} along these two dimensions. StreamMind uses continuous ingestion and satisfies both Eq.~\eqref{eq:harness_continuous} and Eq.~\eqref{eq:attn_causal}. Every baseline instead reconstructs either a causal prefix or a bounded recent window for each task and carries no hidden video state across questions.

% =============================================================
% Table: Evaluation harness and model design of systems on StreamArena.
%
% Two orthogonal axes:
%   Harness: Prefix replay      = reactive causal-prefix replay; proactive forward monitoring
%            Recent-window      = bounded causal-window reconstruction
%            Continuous         = frames streamed from t=0, no rewind
%   Model:   Non-causal    = bidirectional attention over all frames at once
%            Causal        = prefix-only visual tokens
%              windowed    = bounded visual buffer rolled forward one segment at a time
%              summarized  = rolling textual memory condensing past visuals
%              compressed  = architectural KV/token or reasoning compression
%              multi-worker = decoupled frontend interaction + backend memory/retrieval
% =============================================================
\begin{table*}[t]
\centering
\small
\setlength{\tabcolsep}{4pt}
\renewcommand{\arraystretch}{1.15}
\begin{tabularx}{\linewidth}{@{}>{\raggedright\arraybackslash}p{0.24\linewidth}
    >{\raggedright\arraybackslash}X
    >{\raggedright\arraybackslash}X
    >{\raggedright\arraybackslash}X@{}}
\toprule
\textbf{System} & \textbf{Harness} & \textbf{Model} & \textbf{Design class} \\
\midrule
Offline turn-based MLLMs (A) & Prefix replay & Non-causal & Offline turn-based \\
AURA (B)                   & Recent-window & Causal, windowed    & Recent-window \\
MiniCPM-o-4.5 (B)          & Recent-window & Causal, windowed    & Recent-window \\
VST (C)                    & Prefix replay & Causal, summarized  & Text-summary \\
StreamForest (D)           & Prefix replay & Causal, compressed  & Model-internal compression \\
ThinkStream (D)            & Prefix replay / forward monitoring & Causal, reasoning-compressed & Model-internal compression \\
\midrule
\rowcolor{gray!12}
\textbf{StreamMind (E, Ours)} & \textbf{Continuous} & \textbf{Causal, multi-worker} & \textbf{Persistent multi-worker} \\
\bottomrule
\end{tabularx}
\caption{Evaluation protocol taxonomy. Harness describes input delivery, Model describes temporal processing, and Design class identifies the architecture without conflating it with the evaluated harness. ThinkStream uses causal-prefix replay for reactive queries and forward monitoring for proactive tasks.}
\label{tab:protocol_taxonomy}
\end{table*}

To preserve each method's intended inference mechanism, we adapt its released interface to StreamArena rather than forcing all baselines into a common continuous-ingestion implementation. Each question invokes the corresponding inference routine with the method-specific causal interval, sampling policy, and memory mechanism described below. No reactive query receives frames after $t_i^q$, and multi-turn textual history is preserved within each video. The resulting comparison therefore evaluates the released streaming mechanisms under a common causal, open-ended benchmark while retaining their different context policies.

\subsection{Offline turn-based MLLMs}
\label{app:setup_offline}

All offline turn-based MLLMs in Group~(A) share the following per-question protocol.

\paragraph{Visual input.}
We uniformly sample from $\mathcal{S}_{0:t^{q}_i}$ a frame set of size
\begin{equation}
\label{eq:offline_frame_budget}
    N_{\text{frame}} = \min\!\big(N_{\max},\; \lceil t^{q}_i \rceil\big), \qquad N_{\max} = 128,
\end{equation}
with each frame resized so that no side exceeds $1280 \times 720$. The short-window cap $\lceil t^{q}_i \rceil$ prevents near-duplicate frames when $t^{q}_i < N_{\max}$ seconds.

\paragraph{Audio and subtitle input.}
Let $\mathcal{M}_{\text{omni}} = \{\text{MiMo-V2.5},\text{Qwen3.5-Omni},\text{Gemini~3.5~Flash}\}$ be the omni-capable models, and let $\mathcal{L}=\{\text{zh-CN},\text{zh-Hans},\text{en},\text{zh-Hant}\}$ be the caption-language priority list. For $m \in \mathcal{M}_{\text{omni}}$, we directly provide the raw 16 kHz mono audio prefix $a_{0:t_i^q}$. Otherwise, we provide
\begin{equation}
\label{eq:subtitle_fallback}
    c_{0:t^{q}_i} = 
    \begin{cases}
        \text{YT}(v,\ell),
            & \text{if an }\ell\in\mathcal{L}\text{ track exists}, \\
        \text{ASR}(a_{0:t^{q}_i}),
            & \text{otherwise},
    \end{cases}
\end{equation}
where $\text{YT}(v,\ell)$ returns the native YouTube caption track in language $\ell$, following the order in $\mathcal L$, and $\text{ASR}(\cdot)$ denotes the Qwen3-ASR-1.7B fallback stored as a \texttt{.vtt} file.

\paragraph{Sampling and tool use.}
We use temperature $T=0.6$, top-$p=0.95$, and a generation budget of $K=65{,}536$ tokens for all offline MLLMs. Gemini~3.5~Flash additionally uses high reasoning effort and high media resolution. Tool-utilization tasks share a common tool-calling loop with at most $R_{\max}=5$ rounds; both text and image search are routed through Google.

\subsection{Recent-window and text-summary methods}
\label{app:setup_pseudo_streaming}

These methods reconstruct their designated causal input after a query or monitoring instruction arrives. Between tasks, they neither ingest video nor preserve hidden video state. Their causal intervals and segmentation differ by method and are specified below.

\paragraph{AURA (recent-window).}
For a reactive query, AURA reconstructs at most the latest 30 one-second video chunks ending at $t_i^q$. The chunks form one model request in temporal order: earlier chunks are paired with a silence token, while the final chunk carries the question and all captions available by $t_i^q$. Decoding is deterministic with a budget of 1,024 tokens. For a proactive task, the monitoring instruction is registered at $t_i^q$, after which the harness checks the model once per video second until two seconds after the annotated event. Each check independently reconstructs the latest 30 one-second chunks available at that time and supplies captions only up to the same timestamp. Any response before the event is marked premature; responses at the event time or either of the next two sampled seconds are candidate alerts. Different videos run concurrently, whereas turns within one video remain sequential. The evaluated AURA implementation has no external tool-calling loop, so Tool is reported as N/A.

\paragraph{MiniCPM-o-4.5 (recent-window, unified).}
Let $\mathcal{Q}_{\text{HD}}$ denote reactive queries and $\mathcal{Q}_{\text{FD}}$ proactive queries. MiniCPM-o-4.5 selects its rollout interval as
\begin{equation}
\label{eq:minicpmo_window}
    [\tau_0, \tau_{K_i}] = 
    \begin{cases}
        [\max(0,t^{q}_i-30),t^{q}_i), & i \in \mathcal{Q}_{\text{HD}}, \\
        [t^{q}_i,t^{\text{gt}}_i+w_e), & i \in \mathcal{Q}_{\text{FD}},
    \end{cases}
\end{equation}
with $w_e=3$ seconds. Each second contributes one frame and its corresponding one-second raw-audio segment. Reactive queries use half-duplex generation over the recent window. If the model requests external information, the harness executes Google text search, full-frame image search, or crop-based image search and returns the observations for another generation round, with at most $R_{\max}=5$ rounds. Proactive tasks use the native full-duplex interface with a sliding context of 30 audio-visual units. The model remains silent on $[t_i^q,t_i^{\mathrm{gt}})$; outputs at $t_i^{\mathrm{gt}}$ and the following two sampled seconds are candidate alerts. MiniCPM-o consumes raw audio directly and does not receive caption text.

\paragraph{VST (text-summary).}
For each query, VST samples the available causal prefix at 2 fps and uniformly reduces it to at most 384 frames. Let $d_i=\min(t_i^q,T)$ denote the available prefix duration. VST partitions the sampled prefix into $K_i$ approximately equal temporal segments, where
\begin{equation}
\label{eq:vst_segments}
K_i=
\begin{cases}
2, & d_i\le 30\,\text{s},\\
3, & 30\,\text{s}<d_i<4\,\text{min},\\
5, & d_i\ge4\,\text{min}.
\end{cases}
\end{equation}
The first $K_i-1$ segments are processed sequentially into a textual memory, and the final pass consumes that memory together with the last visual segment, captions available by $t_i^q$, multi-turn dialogue history, and the current question. The visual budget is equivalent to 8,192 tokens per segment; intermediate and final generations are capped at 5,000 and 32,768 tokens, respectively. All state is reconstructed per query and discarded afterward. The evaluated VST implementation supports neither proactive monitoring nor an external tool-calling loop, so Pro and Tool are reported as N/A.

\subsection{Model-internal compression methods}
\label{app:setup_native_streaming}

Model-internal compression methods employ causal compression rather than an explicit rolling text summary. Their verified harness behavior is documented below.

\paragraph{StreamForest.}
For each reactive query, StreamForest reconstructs the causal prefix $\mathcal{S}_{0:t_i^q}$, samples it dynamically at approximately 1 fps with at most 2,048 frames, and applies its hierarchical frame-forest compression in a single generation call. The prompt includes the sampled temporal range, all captions available by $t_i^q$, multi-turn dialogue history, and the current question. Decoding is greedy with a budget of 1,024 tokens. The evaluated implementation supports neither proactive monitoring nor an external tool-calling loop, so Pro and Tool are reported as N/A.

\paragraph{ThinkStream.}
For each reactive query, ThinkStream replays the causal prefix from $t=0$ to $t_i^q$ and allows a 3-second response window after the query. The prefix is processed by the model's streaming-window inference engine using at most 120 chunks of two frames, a context length of 24,576 tokens, a thinking budget of 20 tokens, and an answer budget of 256 tokens. The query includes captions available by $t_i^q$ and the preceding question-answer history from the same video. We use the official response-triggering mechanism and take the first nonempty text emitted after the \texttt{<response>} token. For proactive tasks, the monitoring instruction is injected at $t_i^q$ with no future captions, and the model processes one-second chunks until two seconds after $t_i^{\mathrm{gt}}$. Any response before $t_i^{\mathrm{gt}}$ is premature, while responses at $t_i^{\mathrm{gt}}$ or either of the next two sampled seconds are candidate alerts. The evaluated implementation has no external search loop; its Tool score therefore measures answer correctness without guaranteed tool invocation.

\subsection{StreamMind (ours)}
\label{app:setup_streammind}

\paragraph{Continuous ingestion harness.}
Unlike every baseline above, StreamMind consumes video under Eq.~\eqref{eq:harness_continuous}: the driver decodes frames with OpenCV at $f_{\text{cap}}=2$ fps, pushes them into a ring buffer in temporal order, and never rewinds. For non-proactive queries, it pauses the video clock at $t_i^q$ until the response finishes, preventing inference latency from shifting subsequent timestamps. Proactive tasks do not pause ingestion. In both cases, the model cannot access future frames or replay discarded frames.

\paragraph{Multi-worker architecture.}
StreamMind combines query-driven workers, dynamically spawned Monitor Workers, and a periodic Memory Writer. They share a FrameBuffer $\mathcal{B}_t$, a bounded conversation log $\mathcal{L}_t$, and a Memory Bank
\begin{equation}
    \mathcal{H}_t=\left(\mathcal{G}_t,
    \mathcal{E}_t^{\mathrm{micro}},
    \mathcal{E}_t^{\mathrm{macro}},
    \mathcal{E}_t^{\mathrm{super}},
    \mathcal{F}_t\right),
\end{equation}
where $\mathcal{G}_t$ is a typed entity relation graph and $\mathcal{F}_t$ stores persistent key frames. We report the effective evaluation configuration below, including runtime values that override implementation defaults.

\begin{itemize}[leftmargin=1.4em, itemsep=3pt, topsep=2pt]
    \item \textbf{FrameBuffer and audio observations.} The driver captures at $f_{\mathrm{cap}}=2$ fps, resizes each frame to a maximum side length of 720 pixels, and encodes it as JPEG at quality 75. A $W_{\mathcal B}=60$ s FIFO buffer therefore holds approximately 120 frames. Audio is divided into $W_{\mathrm{env}}=5$ s windows and sent to ASR when its RMS is at least $\rho_{\min}=0.005$. The audio log stores at most 300 observations, and the full conversation log is capped at $10^5$ characters.

    \item \textbf{Front Worker.} On each user utterance, the Front Worker inspects the latest $N_{\mathrm F}=12$ frames at or before $t_i^q$, the conversation log, current facts, two recent macro summaries, and active retrieval briefs. It then answers directly, delegates a retrieval brief, spawns a Monitor, or cancels a Monitor:
    \begin{equation}
    \label{eq:front_decision}
        d_i = f_{\mathrm F}\!\left(q_i,\mathcal{B}_{\le t_i^q},
        \mathcal{H}_{\le t_i^q},\mathcal{L}_{\le t_i^q};
        T_{\mathrm F}=0.3,K_{\mathrm F}=256\right).
    \end{equation}
    The Router normally returns the final answer within its ReAct output. A fallback answer pass uses $T=0.4$, $K=1{,}536$, up to 12 query-time frames, 2 current frames, 16 conversation turns, and at most 4 recalled frames.

    \item \textbf{Monitor Worker.} Each monitoring intent creates an independent task that checks the latest $N_{\mathrm M}=4$ frames every $\Delta_{\mathrm M}=2$ s. Its JSON verdict uses $T_{\mathrm M}=0.2$ and $K_{\mathrm M}=512$, and fires only when confidence reaches $\theta_{\mathrm M}=0.6$. The system supports at most $M_{\max}=8$ active Monitors. One-time Monitors terminate after firing; repeating Monitors enforce a 3-second minimum trigger gap and retain 60 visual trigger records plus 200 textual summaries. The agent-side TTL is disabled, while the evaluation driver ends an active proactive case 30 seconds after its annotated answer time. The trigger time is the latest observed frame timestamp, recorded before notification delivery.

    \item \textbf{Router Worker.} The Router runs at most $R_{\mathrm R}=4$ ReAct rounds with $T_{\mathrm R}=0.2$, $K_{\mathrm R}=2{,}048$, $N_{\mathrm R}=8$ frames, and 16 recent dialogue turns. Each round emits an answer or a set of Search and Recall tasks. All tasks in a round execute concurrently, and Search can use either the current frame or a recalled frame identifier as its visual anchor.

    \item \textbf{Search Worker.} Search runs at most $R_{\mathrm S}=3$ rounds with $T_{\mathrm S}=0.2$, $K_{\mathrm S}=4{,}096$, and $N_{\mathrm S}=30$ frames. The launch script restricts both text and image retrieval to Google. It exposes three tools:
    \begin{itemize}[leftmargin=1.2em, itemsep=1pt, topsep=1pt]
        \item \emph{Text search} returns up to $k=10$ results after filtering at similarity $\theta_{\mathrm{sim}}=0.6$.
        \item \emph{Full-frame image search} performs reverse image search on the anchor frame.
        \item \emph{Crop-based image search} follows the HyperEyes protocol~\cite{li2026hypereyes}: it maps a box $[x_1,y_1,x_2,y_2]$ from a $1000\times1000$ coordinate system, crops the anchor, and limits its longest side to 720 pixels. Without a box, it searches the full frame and center crop concurrently.
    \end{itemize}
    Tools within one round execute concurrently.

    \item \textbf{Recall Worker.} Recall runs an Observe, Reason, Act loop of at most $R_{\mathrm C}=4$ rounds with $T_{\mathrm C}=0.2$, $K_{\mathrm C}=2{,}048$, and $N_{\mathrm C}=4$ decision frames. Every read is bounded by the query time $t_i^q$, so no returned event satisfies $t_{\mathrm{end}}>t_i^q$. Its six exposed tools are:
    \begin{itemize}[leftmargin=1.2em, itemsep=1pt, topsep=1pt]
        \item \emph{Time-window search} retrieves micro-events between $t_s$ and $t_e$.
        \item \emph{Entity search} performs fuzzy matching over entity names, aliases, and attributes, returning $k=5$ candidates by default.
        \item \emph{Micro-event search} matches keywords against micro-event descriptions and returns $k=5$ results.
        \item \emph{Entity-conditioned event retrieval} returns all events involving a selected entity.
        \item \emph{Entity-conditioned frame retrieval} returns all key frames associated with a selected entity.
        \item \emph{Relation traversal} performs multi-hop traversal of the entity relation graph.
    \end{itemize}
    Memory calls in one round execute concurrently. A distillation pass with $K_{\mathrm{distill}}=512$ compresses each result into one to three task-specific sentences. A final visual pass checks up to 8 recalled frames and removes frames that do not show the target; if parsing fails or all frames are rejected, the unfiltered set is retained.

    \item \textbf{Memory Writer.} The launch script sets $\Delta_{\mathrm W}=4$ s, overriding the 5 s class default. Each tick reads $N_{\mathrm W}=30$ frames and the last 15 seconds of ASR, then emits an event-boundary decision, key frames, a micro-event record, entity updates, and relation edges with $T_{\mathrm W}=0.2$ and $K_{\mathrm W}=2{,}560$. A micro event closes at a predicted boundary, after 6 ticks, or after 30 seconds. A macro event closes after 5 micro events or 180 seconds; a super event closes after 4 macro events or 900 seconds. Entity canonicalization uses threshold 0.85. The FrameStore retains at most 4,000 images and deduplicates within an 8-second lookback using a dHash Hamming threshold of 6.
\end{itemize}

\paragraph{Backbone.}
The launch configuration uses a shared Qwen3.5-397B-A17B vLLM endpoint for all workers. The Front Worker uses non-thinking mode, whereas backend workers use thinking mode. The implementation supports separate Front, Memory Writer, and Monitor endpoints, but these overrides are empty in the reported configuration.

\subsection{Computing infrastructure}
\label{app:setup_infrastructure}

All experiments run on the following infrastructure.

\paragraph{Hardware.}
\begin{itemize}[leftmargin=1.4em, itemsep=2pt, topsep=2pt]
    \item GPUs: $8\times$NVIDIA H800 (80\,GB HBM3) per server.
    \item CPU: $2\times$Intel Xeon Platinum 8563C (208 logical cores, 2 NUMA nodes).
    \item System memory: 2\,TB DDR5.
\end{itemize}

\paragraph{Software.}
\begin{itemize}[leftmargin=1.4em, itemsep=2pt, topsep=2pt]
    \item Operating system: Ubuntu 22.04.4 LTS (kernel 4.18.0-348.7.1.el8\_5).
    \item CUDA runtime 12.8, NVIDIA driver 570.124.06.
    \item Python 3.10.20.
    \item Inference engine: vLLM 0.25.0.
    \item Key libraries: PyTorch 2.11.0 (\texttt{+cu129}), transformers 5.5.4, OpenCV 4.13.0.
    \item ASR fallback: \texttt{Qwen/Qwen3-ASR-1.7B}.
    \item Judge and closed-source baseline: Gemini~3.1~Pro (judge, \texttt{temperature=0}) and Gemini~3.5~Flash (baseline), called via their public APIs between 2026-05 and 2026-07; snapshot behavior therefore reflects that time window.
\end{itemize}

\paragraph{Judging protocol.}
All open-ended responses are judged by Gemini~3.1~Pro using the factual-equivalence criterion defined in Eq.~\eqref{eq:acc}. This judge is distinct from the evaluated Gemini~3.5~Flash baseline. Every system is scored under the same judge, the same proactive timing rule in Eq.~\eqref{eq:pro_time_window}, and the same random seed for offline frame sampling.

\subsection{Reproducibility artifacts}
\label{app:reproducibility_artifacts}

This subsection documents the artifacts released with the paper and the practical settings needed to reproduce the reported numbers. It complements the AAAI Reproducibility Checklist.

\paragraph{Data release.}
StreamArena is released together with the paper. To respect YouTube's terms of service, we do not redistribute the raw video files. Instead, each of the 243 videos is identified by its YouTube video ID, and every question is stored as a structured JSON record. Each record contains the following fields:

\begin{itemize}[leftmargin=1.4em, itemsep=2pt, topsep=2pt]
    \item \texttt{video\_id}: YouTube 11-character identifier.
    \item \texttt{domain} and \texttt{l2\_subdomain}: coarse and fine category (Section~\ref{app:statistics_full}).
    \item \texttt{capability}: one of \{RTP, HR, Tool, Pro\}.
    \item \texttt{question}: the natural-language query.
    \item \texttt{query\_time}: query timestamp $t^{q}_i$ in seconds.
    \item \texttt{answer}: the reference answer $a^*_i$.
    \item \texttt{evidence}: a list of $\{t_{\text{start}}, t_{\text{end}}, \text{description}\}$ triples; for RTP a single short window, for HR one or more past segments with $t < t^{q}_i$, for Pro the annotated trigger time $t^{\text{gt}}_i$.
    \item \texttt{reasoning\_pattern} (HR only): one of \{single\_point, multi\_count, multi\_recall, multi\_compare, temporal\_range\}.
    \item \texttt{monitoring\_horizon} (Pro only): $t^{\text{gt}}_i - t^{q}_i$ in seconds.
\end{itemize}

\paragraph{Data license and fallback for expired videos.}
StreamArena annotations are released under CC BY 4.0. We track video availability and periodically refresh the index. When a YouTube video ID becomes unreachable, we redirect the corresponding annotations to a compatible archived copy hosted on a public preservation platform, and the release notes list every substitution. We never redistribute the original video content.

\paragraph{Code release.}
The evaluation harness (frame sampling, ASR fallback, subtitle extraction, judge invocation), all baseline adapters used in Table~\ref{tab:main_results}, and the full StreamMind implementation (Front, Monitor, Router, Search, Recall, and Memory Writer workers) are released under Apache-2.0 through the public code repository linked on the first page.

\paragraph{Code-to-paper mapping.}
Every StreamMind worker module includes a docstring that maps its implementation role to the relevant hyperparameters documented in Appendix~\ref{app:setup_streammind}. Data-processing scripts (video downloading via public YouTube IDs, subtitle fetching with the priority list of Eq.~\eqref{eq:subtitle_fallback}, ASR fallback, and 2\,fps frame extraction) live under \texttt{tools/preprocess/} and mirror the description in Section~\ref{app:setup_offline} and Appendix~\ref{app:setup_streammind}.

\paragraph{Randomness and seeds.}
All non-LLM randomness (offline uniform frame sampling of Eq.~\eqref{eq:offline_frame_budget}, tie-breaking in baseline harnesses, and Monitor scheduling in StreamMind) is controlled by a single global seed set to \texttt{42} through Python's \texttt{random}, \texttt{numpy}, and \texttt{torch} generators. Every local vLLM decoding request also uses seed \texttt{42}, together with the temperature and top-$p$ settings reported in Appendix~\ref{app:setup_offline} and Appendix~\ref{app:setup_streammind}. The Gemini~3.1~Pro judge uses \texttt{temperature=0} and is deterministic up to the API's own snapshot policy.

\paragraph{Hyperparameter selection.}
We did not perform automated hyperparameter search. All StreamMind worker settings (temperatures, top-$p$, context frame budgets, token budgets, retrieval round caps, monitor thresholds, and Memory Writer intervals) reported in Appendix~\ref{app:setup_streammind} are the final values used to produce every number in Table~\ref{tab:main_results} and Table~\ref{tab:latency_comparison}. The diagnostic sweeps over frame count (4--200), spatial resolution (240p--720p), and thinking mode (Figure~\ref{fig:diagnostic_ablation}) are ablations rather than tuning, and their settings are not used to select final values.

\paragraph{Number of runs.}
Each cell in Table~\ref{tab:main_results}, Table~\ref{tab:latency_comparison}, Table~\ref{tab:subset_modality}, and Figure~\ref{fig:diagnostic_ablation} corresponds to a single evaluation run. Because a full StreamArena pass covers 3{,}646 questions over 243 hour-scale videos, and StreamMind and several baselines rely on a shared 397B-parameter backbone served through vLLM, repeated runs are not economical. We therefore do not report standard deviations or statistical significance tests, and note this as a limitation of the current evaluation.

\end{document}